# Relation Detection for Indonesian Language using Deep Neural Network - Support Vector Machine

Ramos Janoah Hasudungan
PT. Prosa Solusi Cerdas
Bandung, Indonesia
ramos.janoah@prosa.ai

Ayu Purwarianti
School of Electrical Engineering and Informatics
Institut Teknologi Bandung
ayu@stei.itb.ac.id

*Abstract*— Relation Detection is a task to determine whether two entities are related or not. In this paper, we employ neural network to do relation detection between two named entities for Indonesian Language. We used feature such as word embedding, position embedding, POS-Tag embedding, and character embedding. For the model, we divide the model into two parts: Front-part classifier (Convolutional layer or LSTM layer) and Back-part classifier (Dense layer or SVM). We did grid search method of neural network hyper parameter and SVM. We used 6000 Indonesian sentences for training process and 1,125 for testing. The best result is 0.8083 on F1-Score using Convolutional Layer as front-part and SVM as back-part.

*Keywords— Relation detection, neural network, SVM*

## I. Introduction and Related Works

The relation detection here is defined as a task to classify whether there is a relation between two named entities or not. Different with common relation extraction where each relation has its label name, we aim to conduct an open relation detection system with limited resource of Indonesian language processing tool. Furthermore, the relation detection is employed as a part of entity database building with dashboard where user can see a set of named entity related with a certain named entity.

There have been several researchers in defining relation between named entities, whether for English, Indonesian or other languages. The approaches are divided into rule based and statistical based. In rule based approach [1][3], rules are defined manually whether using only lexical pattern or using also syntactic pattern. The weakness of rule based approach is the coverage. It is not easy to define all rules manually to handle all possible cases.

For the statistical based approach, the techniques can be supervised learning [1][3][4][5], unsupervised learning [8], and semi-supervised learning[8]. Several of these techniques have employed deep learning technique which is shown got higher accuracy compared to other machine learning technique, especially for a large training data set. The best accuracy was achieved using supervised learning approach. Using supervised learning approaches is basically classifying the relation between two entities in the sentences, and the process become similar with sentence or document classification. The process consist of feature extraction and learning. For example, the feature used by Zhou et al. [8], surveyed by [11] is word based feature, base phrase chunking feature, and external feature. Since we aim to handle the limited resource of language processing tool, we chose to adopt [1] approach that conducted relation extraction using CNN with input layer of word embedding vector and position vector. The main difference is that rather than using dense layer, we employ SVM (Support Vector Machine) as one of best machine learning algorithm in text classification. We also add character embedding as another easily input feature. Since Indonesian POS tagger is already available, we add POS tag embedding in the input layer.

For Indonesian language, there are several researches about relation extraction [10][2][5]. [10] employed rule based relation extraction with lexical information as the input features. [2] employed rule based with constituent parser as the input features. [5] employed statistical based relation extraction with dependency parser result as the input features. None of these researches have conducted deep learning approach to detect the relation between named entities. In our knowledge, this is the first attempt on using deep neural network for relation detection on Indonesian language. Here, we try to solve relation detection using deep neural network, combined with SVM in the output layer.

This paper consists of 4 sections including this section as first section. In section two, we describe the proposed method on our relation detection. In section three, we explain about data composition, work flow of the experiment, and the result of the experiment.

## II. Indonesian Relation Detection using Deep Neural Network - SVM

Such as mentioned before, we employ deep neural network algorithm and SVM to identity whether two named entities have relation or not. Basically, our system consists of feature extraction, front-part classifier, and back-part classifier. In order to get the best result, we compare several input features and topologies in each part of the system. Each component of the relation detection is described below.

*A. Feature Extraction*

We employ several input features such as below.

 *1) Word Embedding*

We use pre-trained word embedding model using library word2vec. Word embedding model was built using 24 million of Indonesian news. The dimension of the vector that produces from this word embedding model is 100.

 *2) Position embedding*

Indonesian language has a few common with English, such as similar POS-Tag, existence of passive and active form, etc. From this similarity, we found that position embedding that [1] did can also be used, so we extract position embedding feature. Position embedding is similar with word embedding which has index of position as it inputs.

The position index is the relative position of each word compare to first entity and second entity. We used 50 as an output dimension of position embedding, as [1] also did.

*3) POS-tag Embedding*

Since there is an available Indonesian POS Tagger, we try to use POS-Tag embedding as an additional input feature. POS tag is attached to each word and the input embedding layer is then be changed during training (trainable). The dimension of POS-Tag embedding is a parameter to be grid-searched on experiment.

After all the features are extracted, the features would be concatenated. Overall process is illustrated by Fig. 1.

Besides word, position, and POS-Tag embedding, we also conduct experiment on using character embedding. We implement the character embedding by using convolutional and max-pool layer.

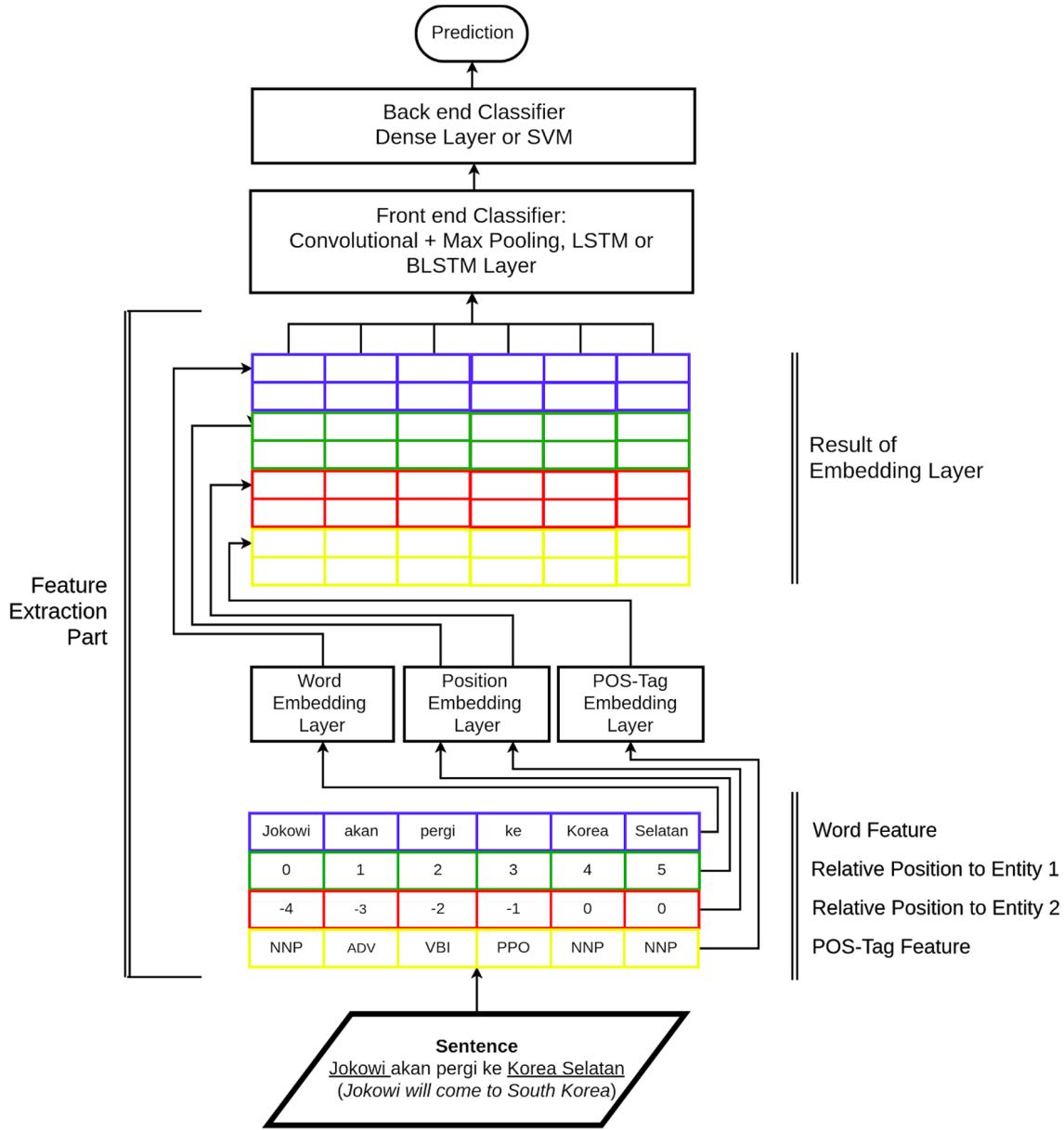

Fig. 1. The overall process of proposed method. The first part is feature extraction part, continue to front-part classifier and ended at back-part classifier.

*B. Deep Neural Network for front-part classifier*

The features extracted from input sentence are then processed by the deep neural network part. We chose to conduct the experiments for CNN, LSTM, and Bi-LSTM topologies, with drop out layer.

*a) Convolutional Neural Network (CNN)*

In this part, we adapt the method from [1] to do the classification, with different amount filter and window size. After convolutional process, we used max pooling layer, and then connect to the back-part of classifier. Illustration of the Convolutional network can be seen on Fig. 2.

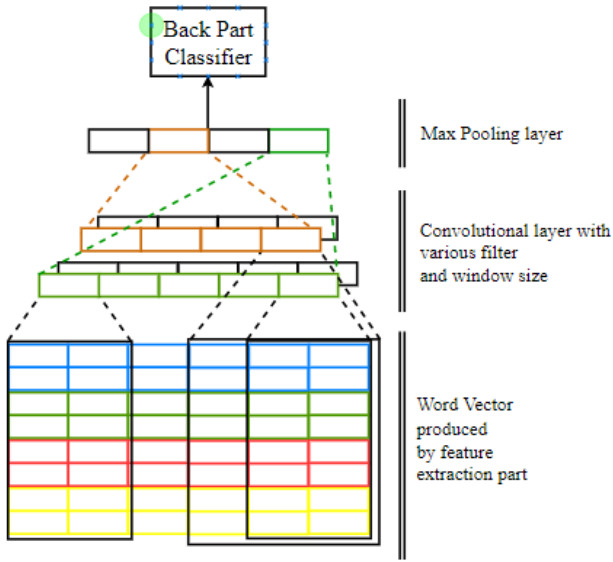

Fig. 2. Used Topologies for Convolutional Neural Network.

*2) LSTM and Bi-LSTM*

In this part, we apply method used by [12] for LSTM and Bi-LSTM part. LSTM is one of Recurrent Neural Network that has ability to learn from sequences, and it has been used for many natural language processing text part, especially for text classification [12][13][14]. Illustration of LSTM and Bi-LSTM can be seen on Fig. 3 and Fig. 4.

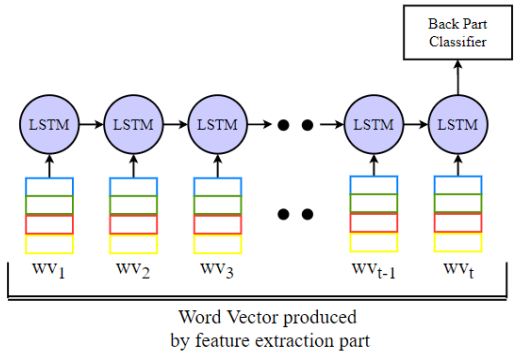

Fig. 3. Used Topologies for LSTM.

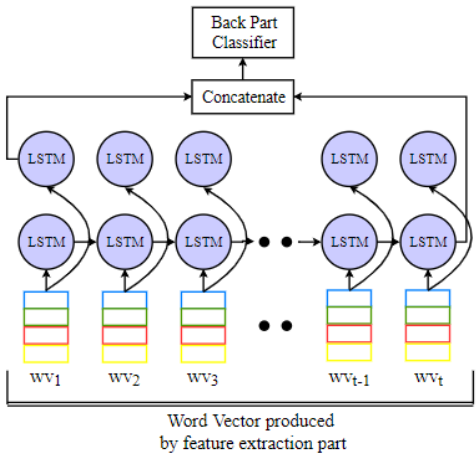

Fig. 4. Used Topologies for Bidirectional LSTM

*C. Dense layer or SVM as back-part of classifier*

For the back-part of classifier, we compare dense layer and support vector machine (SVM). We found that in several researches, SVM had outperformed neural network or multilayer perceptron. If we used the SVM, the dense layer still be used for training the neural network part. After the neural network trained and dense, the output of drop out layer would be the input vector for SVM. We produced the output vector for dropout for all training instances, and the we trained the SVM from that output vector.

### III. EXPERIMENT AND DISCUSSION

*A. Experimental Data*

For the experiment, we used 6,000 sentences as the training data, consists of 14,029 instances of true relation label and 16,019 instances of false relation label. As for the testing, we used 1,100 sentences, with 5,431 instances of true relation label and 7,198 instances of false relation label. The example of the data is illustrated in Fig. 5.

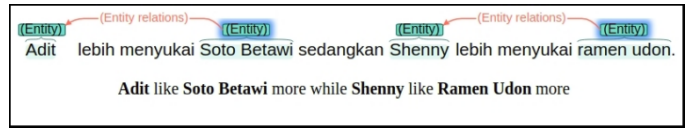

Fig. 5. Indonesian Relation Detection Data Example.

*B. Experimental Result*

We conducted several experiments with aim to get the highest result by comparing parameter combination, input features and neural network topology. For the first experiment, we investigated best parameters for each deep neural network topology. The parameter values compared in the experiment are shown in Table I, and fix parameter on Table II.

TABLE I. PARAMETER VALUES FOR DEEP NEURAL NETWORK OF INDONESIAN RELATION DETECTION

| Front-NN | Parameter | Possible Value |
|---|---|---|
| CNN | Filter | 50, 100, 150, 200 |
|  | Window | [2, 3], [2, 3, 4], [2, 3, 4, 5] |
|  | Dropout | 0.25, 0.50, 0.75 |
|  | POS-Tag Dim | 0, 50, 100 |
| LSTM | Dropout | 0, 0.25, 0.50 |
|  | Softamax Unit | 50, 100, 150, 200 |
|  | POS-Tag Dim | 0, 50, 100 |
| Bi-LSTM | Dropout | 0, 0.25, 0.50 |
|  | Softamax Unit | 50, 100, 150, 200 |
|  | POS-Tag Dim | 0, 50, 100 |

TABLE II. FIXED PARAMETER ON GRID SEARCH STEP

| Configuration | Value of Configuration |
|---|---|
| Training Epoch | 2 |
| Activation | tan h |
| Dense Activation | softmax |
| Position Output Dim | 50 |
| Optimizer | adam |

TABLE III. BEST EXPERIMENT RESULT OF EACH DEEP NEURAL NETWORK TOPOLOGY USED IN FRONT NEURAL NETWORK FOR INDONESIAN RELATION DETECTION WITH WORD EMBEDDING, POSITION EMBEDDING AND POS TAG EMBEDDING

| Front-NN (Best result only) | F1 | P-True | R-True | P-False | R-False |
|---|---|---|---|---|---|
| CNN (Filter=150; Window=2,3,4; Dropout=0.25; No POS Tag) | 0,7745 | 0,7498 | 0,7163 | 0,7929 | 0,8197 |
| CNN (Filter=150; Window=2,3,4; Dropout=0.25; POS Tag dimension=50) | 0,7864 | 0,7538 | 0,7479 | 0,8109 | 0,8156 |
| LSTM (Dropout=0.25; Dense unit=50; POS Tag dimension=50 | 0,4138 | 0,0000 | 0,0000 | 0,5700 | 1,0000 |
| Bi-LSTM (Dropout=0; Dense unit=150; No POS Tag) | 0,7259 | 0,7181 | 0,0789 | 0,7954 | 0,7620 |
| Bi-LSTM (Dropout=0; Dense unit=200; POS Tag) | **0,7880** | 0,7604 | 0,7415 | 0,8085 | 0,8327 |

Table III shows that LSTM model failed to predict 'FALSE' relation between named entity for all instances. Here, we only did comparison on the POS Tag embedding, since word embedding and position embedding are the must-exist input features for our approach. For the POS Tag embedding, the experimental result on Bi-LSTM and CNN showed that the POS Tag embedding can improve the F1-score. The best result was achieved by using Bi-LSTM with no drop out and 200 dense unit.

As has been mentioned before, we also investigated character embedding feature as an easily prepared feature and might be able to handle OOV problems. We used best parameter configuration taken from Table III. Using only CNN and bi-LSTM topologies, the experimental results on character embedding are shown in Table IV. The results showed that using character embedding gave lower F1-score compared to the one without character embedding.

TABLE IV. EXPERIMENTAL RESULT ON INDONESIAN RELATION DETECTION WITH CHARACTER EMBEDDING FOR CNN AND BI-LSTM TOPOLOGIES

| Front-NN | Char Embedding | F1 | P-True | R-True | P-False | R-False |
|---|---|---|---|---|---|---|
| CNN | TRUE | 0.7678 | 0,7944 | 0,6352 | 0,7609 | 0,8759 |
| CNN | FALSE | 0.7864 | 0,7538 | 0,7479 | 0.8109 | 0.8156 |
| Bi-LSTM | TRUE | 0.7725 | 0,7397 | 0,7275 | 0,7969 | 0.8069 |
| Bi-LSTM | FALSE | **0.7880** | 0,7604 | 0,7415 | 0,8085 | 0,8327 |

The last experiment in our research was to compare the dense layer and SVM in the output layer. For the SVM, we've tried several parameters. We implied radial basis function as the kernel, and we grid-searched the parameter C and Gamma, with range $10^{-3}$ until $10^{6}$. Here, we only showed the best parameter for the SVM. The best result of using SVM can be seen on Table V.

TABLE V. COMPARISON OF THE EXPERIMENTAL RESULT ON INDONESIAN RELATION DETECTION USING DENSE LAYER AND SVM FOR CNN AND BI-LSTM TOPOLOGIES

| Topology | F1 | P-True | R-True | P-False | R-False |
|---|---|---|---|---|---|
| CNN +Dense layer | 0.7864 | 0.7538 | 0.7479 | 0.8109 | 0.8156 |
| CNN + SVM (C: $10^0$; Gamma: $10^{-1}$) | **0.8083** | 0.7986 | 0.7441 | 0.8164 | 0.8584 |
| Bi-LSTM + Dense Layer | 0.7880 | 0.7604 | 0.7415 | 0.8085 | 0.8327 |
| Bi-LSTM + SVM (C: $10^5$; Gamma: $10^{-2}$) | 0.7897 | 0.7693 | 0.7321 | 0.8050 | 0.8344 |

## C. Analysis of Experimental Result

To understand the experimental results, we investigated instances of incorrect result of our relation detection. Several cases are shown in Table VI.

TABLE VI. EXAMPLES OF INCORRECT RELATION DETECTION RESULT

| Sentence | Entities | Prediction | Expected |
|---|---|---|---|
| *Sulawesi Tenggara menawarkan beberapa situs alam yang indah, termasuk Kepulauan Wakatobi, Labengki, dan massif karst Matarombéo.* (Southeast Sulawesi offered some beatiful natural sites, including Wakatobi Islands, Labengki, and massive karst of Matarombéo.) | Sulawesi Tenggara - Wakatobi | TRUE | TRUE |
| | Sulawesi Tenggara - Labengki | **FALSE** | **TRUE** |
| | Sulawesi Tenggara - Matarombeo | TRUE | TRUE |
| | Wakatobi - Labengki | **TRUE** | **FALSE** |
| | Wakatobi - Matarombeo | FALSE | FALSE |
| | Labengki - Matarombeo | FALSE | FALSE |
| *Usai menyelenggarakan pernikahan Sabtu (19/5/2018), Pangeran Harry dan Meghan Markle dikabarkan akan berbulan madu ke Kanada.* (After organizing the wedding on Saturday (19/5/2018), Prince Harry and Meghan Markle are rumored to be honeymooning to Canada) | Sabtu (19/5/2018) - Pangeran Harry | TRUE | TRUE |
| | Sabtu (19/5/2018) - Meghan Markle | **FALSE** | **TRUE** |
| | Sabtu (19/5/2018) - Kanada | FALSE | FALSE |
| | Pangeran Harry - Meghan Markle | TRUE | TRUE |
| | Pangeran Harry - Kanada | **FALSE** | **TRUE** |
| | Meghan Markle - Kanada | TRUE | TRUE |

Our first analysis is on the position embedding feature. Table VI shows that nearby entities will have high probability to be classified as related, and far entities will have high probability to be classified as unrelated. This case is shown in fourth instances in Table VI. Thus, this will be an argument to have further research on using sentence parser as an input for the deep neural network.

We also observed that the order of named entity affect the relation detection result. We assumed that this is due to POS tag information of NNP class. A word with NNP reside in the middle of compared entities might cause the entities to be unrelated. This case is shown in 2nd, 8th, and 11th instance in Table VI. This condition strengthens the argument on the usage of sentence parser.

As for the usage of SVM, compared to dense layer, we found that the SVM outperformed the dense layer in all sentences that we found. The examples are shown in Table VII.

TABLE VII. Examples of Relation Detection Result (SVM vs Dense Layer)

| Sentence | Entities | SVM | Dense | Expected |
|---|---|---|---|---|
| *Ribuan warga Kabupaten Biak Numfor, Papua melakukan aksi lanjutan penyalaan seribu lilin menuntut pembebasan Gubernur DKI Jakarta non-aktif Basuki Tjahaja Purnama alias Ahok dalam kasus penistaan agama di lapangan remaja Biak, Sabtu malam.* (Thousands of civilian of Biak Numfor Regency, Papua, carried out further actions to ignite a thousand candles demanding the release of the non-active Jakarta Governor Basuki Tjahaja Purnama alias Ahok in a blasphemy case at the Biak youth court on Saturday night.) | Biak Numfor - Papua | TRUE | TRUE | TRUE |
| | Papua - Gubernur DKI Jakarta | **FALSE** | **TRUE** | **FALSE** |
| | Papua - Basuki Tjahaja Purnama | FALSE | FALSE | FALSE |
| | Gubernur DKI Jakarta - Basuki Tjahaja Purnama | **TRUE** | **FALSE** | **TRUE** |
| *"The Underdogs" bercerita tentang kuartet Ellie (Sheryl Sheinafia), Dio (Brandon Salim), Nanoy (Babe Cabita) dan Bobi (Jeff Smith) yang tersisihkan dan tidak dianggap bertekad mengubah nasib dengan cara menjadi YouTuber.* ("The Underdogs" tells the story of the quartet of Ellie (Sheryl Sheinafia), Dio (Brandon Salim), Nanoy (Babe Cabita) and Bobi (Jeff Smith) who are marginalized and not considered determined to change their fate by becoming a YouTuber.) | The Underdogs - Nanoy (Babe Cabita) | TRUE | TRUE | TRUE |
| | The Underdogs - Bobi (Jeff Smith) | **TRUE** | **FALSE** | **TRUE** |
| | Ellie (Sheryl Sheinafia) - Dio (Brandon) | **FALSE** | **TRUE** | **FALSE** |
| | Ellie (Sheryl Sheinafia) - Bobi (Jeff Smith) | FALSE | FALSE | FALSE |

We also try to see how the mapping of the C and Gamma parameter, so we know which parameter get the most influence of performance.

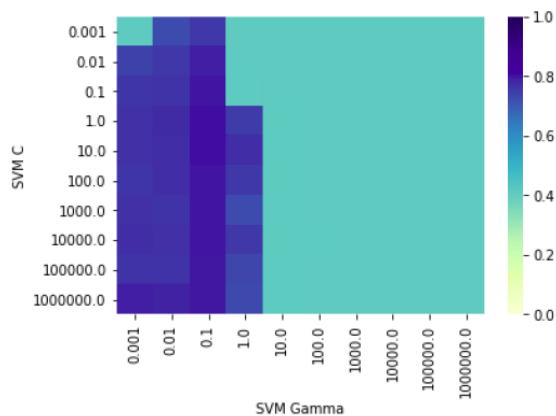

Fig. 6. Heatmap diagram of SVM performance with CNN as front-part classifier. The parameter C is on x-axis and Gamma is on y-axis

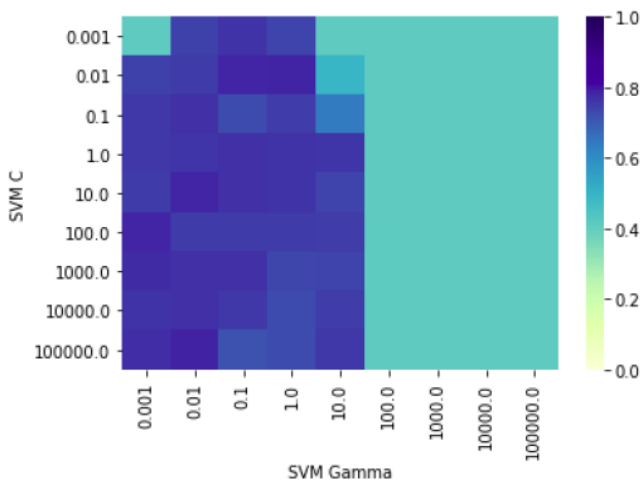

Fig. 7. Heatmap diagram of SVM performance with Bi-LSTM as front part classifier. The parameter C is on x-axis and Gamma is on y-axis

From the heatmap chart from Fig. 6 and Fig. 7, we see that parameter Gamma is influenced more than parameter C. The best range for parameter C is $10^{-3}$ until $10^0$ and $10^{-3}$ until $10^1$ on CNN and Bi-LSTM respectively.

## IV. CONCLUSION

In this paper, we tried to detect relation between named entities on an Indonesian sentence. We implemented different techniques from [1] such as POS-Tag embedding, character embedding feature, LSTM and Bi-LSTM for front-part classifier, SVM as back-part classifier. The best result was 0.8083 using word, position, and POS-Tag embedding, CNN for the front-part classifier, and SVM for back-part classifier. There are so much room for improvement for this task. For example, we can use different structural feature such as dependency parse result of a sentence, to handle several problems caused by position embedding and POS-Tag embedding.

## V. ACKNOWLEDGEMENT

This work was funded by the Indonesian research program 'PENELITIAN TERAPAN UNGGULAN PERGURUAN TINGGI' with title 'Sistem Cerdas Pemantau Perilaku Penggunaan Gadget di Kalangan Remaja Menggunakan Teknik Pembelajaran Mesin' (intelligent system for monitoring gadget usage behavior among teenagers using machine learning). We also would like to thank you other parties who contributed in this experiment or during the process of working for this article.